\documentclass[10pt,twocolumn,letterpaper]{article}

\usepackage{cvpr}              

\definecolor{cvprblue}{rgb}{0.21,0.49,0.74}
\usepackage[pagebackref,breaklinks,colorlinks,allcolors=cvprblue]{hyperref}
\usepackage{booktabs}
\usepackage{multirow}

\usepackage{caption}   

\def\paperID{36674} 
\def\confName{CVPR}
\def\confYear{2026}

\title{Concept-Guided Fine-Tuning: Steering ViTs away from Spurious Correlations to Improve Robustness}

\author{Yehonatan Elisha\\
Tel Aviv University\\
\and
Oren Barkan\\
The Open University\\
\and
Noam Koenigstein\\
Tel Aviv University\\
}

\begin{document}
\maketitle
\begin{abstract}
Vision Transformers (ViTs) often degrade under distribution shifts because they rely on spurious correlations, such as background cues, rather than semantically meaningful features. Existing regularization methods, typically relying on simple foreground-background masks, which fail to capture the fine-grained semantic concepts that define an object (e.g., ``long beak'' and ``wings'' for a ``bird''). As a result, these methods provide limited robustness to distribution shifts. To address this limitation, we introduce a novel finetuning framework that steers model reasoning toward concept-level semantics. Our approach optimizes the model's internal relevance maps to align with spatially grounded concept masks. These masks are generated automatically, without manual annotation: class-relevant concepts are first proposed using an LLM-based, label-free method, and then segmented using a VLM. The finetuning objective aligns relevance with these concept regions while simultaneously suppressing focus on spurious background areas. Extensive experiments on five out-of-distribution benchmarks demonstrate that our method improves robustness across multiple ViT-based models. Furthermore, we show that the resulting relevance maps exhibit stronger alignment with semantic object parts, offering a scalable path toward more robust and interpretable vision models. Finally, we confirm that concept-guided masks provide more effective supervision for model robustness than conventional segmentation maps, supporting our central hypothesis. Our code is provided at: \url{https://github.com/yonisGit/cft}
\end{abstract}    
\section{Introduction}
\label{sec:intro}
\begin{figure}[t]
  \centering
  \includegraphics[width=0.35\textwidth]{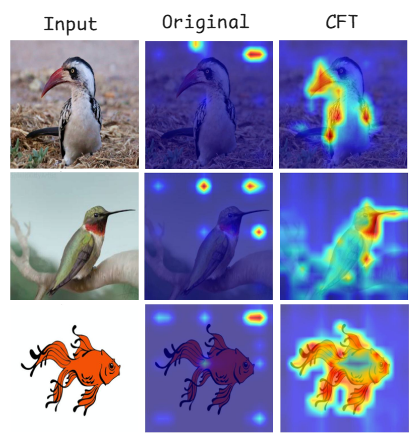}
    \caption{Motivation for CFT: Relevance maps produced by ViTs often concentrate on spurious background cues rather than semantically meaningful concepts. The figure illustrates this issue using ViT-B on ImageNet-A and ImageNet-R, showing relevance maps before and after applying CFT. By encouraging the model to focus on class-relevant, discriminative concepts, CFT substantially improves the semantic alignment of relevance maps. Notably, after CFT, the model highlights meaningful object parts, such as the beak and wings of the bird (top row) or the fins and mouth of the fish (bottom row), despite never being fine-tuned on these datasets.}
  \label{fig:salient_problem}
\end{figure}

Modern ViTs~\cite{dosovitskiy2020image,he2022masked_mae} achieve remarkable performance on standard benchmarks like ImageNet~\cite{deng2009imagenet}, yet their robustness under distribution shifts remains limited. A growing body of evidence shows that these models often rely on spurious correlations, such as background textures or contextual cues, rather than the semantic content of the target object~\cite{geirhos2020shortcut, hendrycks2021many}. This reliance manifests as catastrophic failures on out-of-distribution (OOD) data, including natural adversarial examples~\cite{hendrycks2021natural}, images with altered viewpoints~\cite{barbu2019objectnet}, or artistic renditions~\cite{wang2019learning}. While such behavior may be sufficient for in-distribution accuracy, it undermines trustworthiness in real-world deployment, where environmental conditions are rarely controlled.

\noindent A promising avenue to improve robustness is to align the model's internal reasoning with semantically meaningful image regions. Prior work has shown that models relying on object foregrounds exhibit better generalization under distribution shifts, introducing methods that leverage ground-truth object segmentation masks, for example, by guiding data augmentation strategies~\cite{singh2020dont} or by informing the design of architectural components~\cite{zhu2019deformable}. However, existing approaches either require extensive retraining or annotated ground-truth segmentation masks. These limitations hinder scalability and practical adoption, especially for large pretrained models where fine-tuning must be both efficient and effective. Moreover, binary foreground–background separation can often be too coarse to support robust recognition, as it treats the foreground as a uniform region and overlooks its internal semantic structure. Consider recognizing a ``bird'': robust models should attend to discriminative parts like ``wings" and ``long beak" (top row of Fig.~\ref{fig:salient_problem}) rather than the entire silhouette. Similarly, relevant features may extend beyond the primary object - a "branch" can provide contextual evidence for ``lorikeet", while ``water" can support ``duck" recognition. 

\noindent In this work, we introduce \emph{Concept-Guided Fine-Tuning} (\textbf{CFT}), a post-hoc framework that steers ViTs toward semantically meaningful reasoning without requiring ground-truth masks or full retraining. CFT operates in three stages. First, an LLM-based, label-free method~\cite{oikarinen2023label} proposes a set of 
context-aware semantic concepts per class. Second, a vision-language grounding model (GroundedSAM~\cite{ren2024groundingsam}) spatially localizes these concepts in each training image, producing an adaptive guidance mask. Third, the model is optimized by 
aligning its relevance map, computed via the transformer-faithful AttnLRP method~\cite{achtibat2024attnlrp}, with this concept-based mask, encouraging high relevance within concept regions while suppressing spurious background cues. A 
concurrent classification-consistency objective ensures classification accuracy is preserved throughout fine-tuning. Following the protocol from~\cite{chefer2022optimizing}, we train on half of ImageNet-1K classes, amounting to only 1,500 images (three per class for half the ImageNet-1K classes) with no manual annotation.
Despite this minimal supervision, CFT consistently improves robustness across five OOD benchmarks and three ViT-based models while largely maintaining, and in some cases improving, in-distribution accuracy. The resulting relevance maps exhibit significantly stronger alignment with ground-truth object masks, and robustness gains generalize to held-out classes unseen 
during fine-tuning, confirming that CFT refines the model's underlying reasoning rather than memorizing class-specific cues. Taken together, CFT represents a step toward vision models that are both more robust and more interpretable.

\section{Related Work}
\label{sec:related}
\noindent\textbf{Robustness and Shortcut Learning.}
A primary challenge for modern vision models is their tendency to learn shortcuts, spurious correlations in the training data, such as background textures, that do not generalize~\cite{geirhos2020shortcut}. This reliance limits model robustness on out-of-distribution (OOD) data. Consequently, a suite of challenging benchmarks has been developed to measure this vulnerability, including datasets with natural adversarial examples (ImageNet-A~\cite{hendrycks2021natural}), novel viewpoints and contexts (ObjectNet~\cite{barbu2019objectnet}), artistic renditions (ImageNet-R~\cite{hendrycks2021many}), sketches (ImageNet-Sketch~\cite{wang2019learning}), and synthetic transformations (SI-Score~\cite{djolonga2021robustness}). Model performance is typically contrasted with in-distribution accuracy on standard benchmarks like ImageNet~\cite{deng2009imagenet, russakovsky2015imagenet} and its variants (ImageNet-v2~\cite{recht2019imagenet}). Our work evaluates extensively on these OOD datasets to demonstrate meaningful improvements in robustness.

\noindent\textbf{Saliency-Guided Model Regularization.}
One prominent approach to mitigate shortcut learning is to explicitly guide the model's reasoning. This is often achieved by regularizing the model's explanations to focus on pre-defined foreground regions. For example, Right for the Right Reasons (RRR)~\cite{ross2017right} constrains model explanations to match annotated foreground regions via an input-gradient regularizer. GradMask~\cite{simpson2019gradmask} uses saliency-based gradient masking during backpropagation to reduce overfitting, and RRDA~\cite{Santos_2023_ICCV} employs data augmentation strategies guided by explanation methods to preserve foreground relevance. Many of these approaches act as regularizers during computationally expensive full-training cycles~\cite{ross2017right, viviano2019saliency, Santos_2023_ICCV}, and rely on input gradients that are often unstable or unfaithful proxies for explanations in ViTs~\cite{liu2022rethinking, naseer2021intriguing}. Recent works have explored regularizing Vision Transformers by optimizing their relevance maps. 
Notably, RobustViT~\cite{chefer2022optimizing} fine-tunes ViTs using uniform foreground-background segmentation masks to enhance robustness. Diverging from this reliance on coarse spatial masking, our work leverages concept-level information to provide fine-grained guidance, actively steering the model toward semantically meaningful features.\\
\noindent\textbf{Vision Models Explainability.} Explainable AI has advanced rapidly in recent years, with significant developments across multiple modalities~\cite{elisha2025rethinking,barkan2024llm,barkan2024counterfactual,elisha2024probabilistic,iia,barkan2020explainable,haddad2025sloc,barkan2025fidelity,gurevitch2025lxr,fong2019understanding}. Explainability methods aim to reveal the reasoning behind model predictions. In vision models, relevance maps can highlight the regions that influence a classifier’s decision, and may expose cases where the model overlooks salient features (see Fig.~\ref{fig:salient_problem}, middle column). A dominant family of interpretation methods relies on gradients~\cite{iia,simonyan2013deep,mahendran2016visualizing,dabkowski2017real,elisha2025rethinking,barkan2023stochastic,barkan2021gam,barkan2021grad,elisha2024probabilistic,barkan2025bee}, which have been refined by incorporating additional input signals~\cite{gu2018recent,shrikumar2017learning,smilkov2017smoothgrad,srinivas2019full}. Other prominent approaches include permutation-based techniques grounded in Shapley values~\cite{lundberg2017unified,shrikumar2017learning} and theory-driven attribution propagation methods, such as Layer-wise Relevance Propagation (LRP)~\cite{montavon2017explaining,bach2015pixel}, which propagates the output prediction backward through the network. When applied to transformer architectures, initial work demonstrated that combining gradients and attention values can yield viable interpretations~\cite{chefer2021transformer,barkan2023deep}. Yet, the technical limitations of purely gradient-based explanations for ViTs have motivated the development of more faithful, propagation-based alternatives. AttnLRP~\cite{achtibat2024attnlrp} specifically adapts the LRP principle for transformers by properly attributing relevance through the integration of information from both the attention and MLP blocks. This approach yields stable and faithful relevance maps that are better suited for model refinement than raw gradient-based signals. Consequently, the demonstrated stability and faithfulness~\cite{achtibat2024attnlrp} of AttnLRP make it the clear choice for the explanation backbone of our fine-tuning framework. This choice is further supported by empirical comparisons with alternative saliency methods, provided in the Appendix.\\
\noindent\textbf{Semantic Guidance from Vision-Language Models.}
The bottleneck of requiring human-annotated masks is being rapidly obviated by the rise of powerful vision-language models (VLMs)~\cite{radford2021learning}. Models like Grounding DINO~\cite{liu2023grounding} and Segment Anything (SAM)~\cite{kirillov2023segment}, combined in tools like GroundedSAM~\cite{ren2024groundingsam}, can segment arbitrary semantic concepts in a zero-shot manner from text prompts. This technology unlocks the ability to generate dynamic, concept-level guidance maps, moving decisively beyond the insufficient static foreground-background dichotomy. While prior work has used VLMs for tasks like pseudo-labeling~\cite{zhou2022conditional} or data augmentation~\cite{jia2022visual}, their use as a supervisory signal for spatially grounding a model's internal explanations with specific concepts remains unexplored. 

\section{Method}
\label{sec:method}
We propose Concept-Guided Fine-Tuning (\noindent\textbf{CFT}), a framework for improving the robustness of ViTs by steering their internal reasoning toward semantically meaningful concepts. 
CFT demonstrates the efficiency of concept-guidance through the RobustViT framework~\cite{chefer2022optimizing}.
Specifically, CFT aligns the model's LRP maps with concept-level regions, encouraging the model to rely on discriminative semantic features rather than spurious correlations~\cite{geirhos2020shortcut}. Although our primary focus is on ViTs, we also provide an alternative implementation for CNNs in Sec.~\ref{subsec:relevance}, with additional evaluations reported in Sec.~\ref{sec:experiments}.

\subsection{Problem Setup}
Let $f_\theta: \mathcal{X} \!\to\! \mathcal{Y}$ be a pretrained ViT, where $\mathcal{X}$ is the input image space and $\mathcal{Y}$ is the label space. The model is defined by its parameters $\theta$. For an input image $I \in \mathcal{X}$, the model produces a prediction $f_\theta(I)$. We can also compute the model's relevance map $\Phi(I;\theta)$, which indicates which parts of the image $I$ were most important for the prediction.

\noindent Given a small finetuning dataset $\mathcal{D} = \{(I_j, y_j)\}_{j=1}^N$, consisting of $N$ image-label pairs, our goal is to find optimal parameters $\theta^*$. These new parameters should align the model's relevance map $\Phi(I;\theta^*)$ with a concept-based semantic mask $S(I)$, without harming classification accuracy.

\noindent This objective is formulated as finding the parameters $\theta^*$ that minimize a total loss $\mathcal{L}$:
\begin{equation}
\theta^* = \arg\min_\theta \, 
\mathbb{E}_{(I,y)\sim\mathcal{D}}
\big[
\mathcal{L}(\theta, I, y)
\big].
\label{eq:objective}
\end{equation}
The total loss $\mathcal{L}$ combines a relevance loss and a classification loss, which are detailed in Section~\ref{subsec:alignment_obj}.

\subsection{Relevance and Semantic Guidance}
\label{subsec:relevance}
\noindent\textbf{Relevance Extraction.}
We compute a patch-level relevance map $\Phi(I; \theta) \in [0,1]^{H \times W}$, where $H$ and $W$ denote the height and width of the ViT patch grid. Relevance is derived using Attention-aware Layer-wise Relevance Propagation (AttnLRP)~\cite{achtibat2024attnlrp}, which backpropagates the class output score through the model. The relevance $\Phi_i^{(\ell-1)}$ of token $i$ at layer $\ell-1$ is computed from the relevance $\Phi_j^{(\ell)}$ of tokens $j$ at the next layer $\ell$ as:
\begin{equation}
\Phi_i^{(\ell-1)} =
\sum_{j}
\frac{A_{ij}^{(\ell)} \Phi_j^{(\ell)}}{\sum_k A_{kj}^{(\ell)} + \epsilon},
\label{eq:lrp}
\end{equation}
where $A_{ij}^{(\ell)}$ denotes the attention weight from token $i$ to token $j$, and $\epsilon$ ensures numerical stability.
For CNNs, we adapt AttnLRP by replacing attention maps with intermediate feature representations, following the approach of Barkan et al.~\cite{iia}, combining activation magnitudes with standard LRP relevance scores. We favor LRP-based methods as they satisfy the conservation property, which guarantees that the total relevance propagated through the network sums to the model's output score. This ensures that relevance maps constitute a faithful redistribution of the prediction signal rather than an arbitrary approximation, making them well-suited as an optimization target in our fine-tuning objective.

\noindent\textbf{Concept Set Creation and Validation.}
For a dataset $\mathcal{D'}$ containing classes $C$ with $P$ examples per class, we extract class-discriminative textual attributes $\xi_c$ for each $c \in C$ using the procedure of~\cite{oikarinen2023label}. This produces linguistically interpretable, class-specific candidate concepts.
To ensure reliability, we apply an automated validation step based on visual grounding. Given an image $I$ with label $l$, we provide its corresponding attribute set $\xi_l$ to GroundedSAM~\cite{ren2024groundingsam}, a zero-shot grounding model combining GroundingDINO~\cite{liu2023grounding} with SAM~\cite{kirillov2023segment}. For each concept $k \in \xi_l$, GroundedSAM returns corresponding segmentation masks when the concept is visually present, and no mask otherwise. Thus, for each image $I$ and concept $k$, we obtain a segment set $\text{Seg}_k(I)$, which is empty when $k$ is absent in the image. Concepts are validated according to two criteria:
(1) \textbf{Occurrence Rate} — the fraction of images in class $c$ where $k$ is detected, i.e., $|\{I \in \mathcal{D'}_c : \text{Seg}_k(x) \neq \emptyset\}| / |\mathcal{D'}_c|$, where ${D'}_c$ is a subset of the $D'$ containing images of class $c$, and
(2) \textbf{Spatial Coverage} — the mean IoU between $\bigcup_k \text{Seg}_k$ and the corresponding class-level segmentation mask, measuring how well concepts visually cover their target class regions. Concepts that fail to meet the occurrence criterion are discarded, yielding a validated set of spatially grounded, frequently occurring concepts per class. Although this validation phase produces higher-quality concept sets, one may alternatively use the initial concept set $\xi_c$ for each class $c$ without validation. As demonstrated in the Appendix, the validation step yields superior results but is not mandatory. This process is performed once prior to the fine-tuning stage.

\noindent\textbf{Semantic Mask Generation.}
For each image $I$, we generate a binary semantic guidance mask $S(I) \in \{0, 1\}^{H \times W}$.
Using the validated concept sets from the previous step, we again employ GroundedSAM~\cite{ren2024groundingsam} to obtain binary segmentation masks $M_k(I)$ for all concepts $k$. If a concept is not present in $I$, $M_k(I)$ is set to zero. The final semantic guidance mask $S(I)$ is formed by applying the maximum operator across all individual concept masks.

\subsection{Training Objective}
\label{subsec:alignment_obj}
The total loss $\mathcal{L}$ consists of two weighted components: an alignment loss $\mathcal{L}_{\text{align}}$ and a classification loss $\mathcal{L}_{\text{cls}}$.

\noindent\textbf{Alignment Loss.}
To align the relevance map $\Phi(I)$ with the semantic mask $S(I)$, we define two complementary terms.

\noindent The first term, $\mathcal{L}_{\text{concept}}$, promotes high attribution within the concept regions by minimizing the following objective over all concept pixels:
\begin{equation}
\mathcal{L}_{\text{concept}} = -\frac{1}{|S|}\sum_{p \,\in\, S} \log \Phi_p(I),
\label{eq:concept_loss}
\end{equation}
where $\Phi_p(I)$ denotes the relevance value at pixel $p$, and $S$ indexes the set of concept pixels where $S(I)=1$. This term drives the attribution values inside the concept mask toward their maximum.

\noindent The second term, $\mathcal{L}_{\text{non-concept}}$, suppresses spurious attribution in background regions:
\begin{equation}
\mathcal{L}_{\text{non-concept}} = -\frac{1}{|\bar{S}|}\sum_{p \,\in\, \bar{S}} \log\bigl(1 - \Phi_p(I)\bigr),
\label{eq:non_concept_loss}
\end{equation}
where $\bar{S}$ indexes all pixels where $S(I)=0$. This term penalizes any residual relevance assigned to non-concept areas.

\noindent The total alignment loss combines these two terms:
\begin{equation}
\mathcal{L}_{\text{align}} = \lambda_{\text{concept}}\,\mathcal{L}_{\text{concept}} + \lambda_{\text{non-concept}}\,\mathcal{L}_{\text{non-concept}}.
\label{eq:relevance_loss}
\end{equation}

\noindent\textbf{Classification Loss.}
In the absence of an explicit regularization objective, fine-tuning drives the model to produce explanations that closely resemble the ground-truth segmentation but at the expense of a severe drop in accuracy. To prevent this collapse, it is essential to introduce an auxiliary loss that constrains the fine-tuned model's output distribution to remain consistent with that of the original model. To achieve this balance, we incorporate a classification-consistency loss, defined as follows:
\begin{equation}
\mathcal{L}_{\text{cls}} = \text{CrossEntropy}\bigl(f_{\theta}(I),\, \arg\max\, f_{\theta}(I)\bigr),
\label{eq:cls_loss}
\end{equation}
where $\arg\max\, f_{\theta}(I)$ denotes the class predicted by the model $f_{\theta}$ for the input image $I$. The loss computes the cross-entropy between the model's output distribution and a one-hot target vector that assigns a probability of 1 to the predicted class. In essence, this objective reinforces the model's confidence in its own predictions by amplifying the probability associated with the predicted class. In Section~\ref{subsec:ablation}, we compare the performance of CFT when using our classification-consistency loss with that achieved using a standard ground-truth cross-entropy loss.

\noindent\textbf{Final Loss.}
The final loss $\mathcal{L}$ is the weighted sum of these two objectives:
\begin{equation}
\mathcal{L} = 
\lambda_{\text{align}}\,\mathcal{L}_{\text{align}} + 
\lambda_{\text{cls}}\,\mathcal{L}_{\text{cls}}.
\label{eq:final_loss}
\end{equation}

\section{Experiments}
\label{sec:experiments}

In what follows, we present a comprehensive experimental evaluation of CFT.
Our implementation and experimental setup for fine-tuning builds upon the open-source codebase introduced by Chefer et al. in RobustViT~\cite{chefer2022optimizing}. Our experiments are designed to answer three research questions:\\ \textbf{(i)} Does CFT improve robustness on real-world and synthetic out-of-distribution benchmarks?\\ \textbf{(ii)} Does CFT produce relevance maps that better align with object foregrounds?\\ \textbf{(iii)} Does the benefit of CFT generalize beyond the fine-tuned classes?\\ We compare CFT against four state-of-the-art baselines that similarly regularize saliency maps during training or fine-tuning: \textbf{GradMask}~\cite{simpson2019gradmask}, Right for the Right Reasons (\textbf{RRR})~\cite{ross2017right}, \textbf{RobustViT}~\cite{chefer2022optimizing}, and \textbf{RRDA}~\cite{Santos_2023_ICCV}. All experiments are conducted on four modern vision models: \textbf{DINOv2}~\cite{oquab2023dinov2}, \textbf{ViT-B}~\cite{dosovitskiy2020image}, DeiT-III (\textbf{DeiT})~\cite{touvron2022deit}, and ConvNeXt-V2 (\textbf{CNv2})~\cite{woo2023convnext}. All models were sourced from the \texttt{timm} library and utilize their corresponding dataset pretrained weights. For DINOv2, we employ the fine-tuned variant designed for image classification.

\subsection{Experimental Setup}

\noindent\textbf{Datasets.}
We evaluate robustness on five standard out-of-distribution benchmarks:
\begin{enumerate}
    \item ImageNet-A (\textbf{IN-A})~\cite{hendrycks2021natural}: a collection of natural adversarial examples where standard ImageNet models fail.
    \item \textbf{ObjectNet}~\cite{barbu2019objectnet}: images with controlled object pose, background, and viewpoint variations.
    \item ImageNet-R (\textbf{IN-R})~\cite{hendrycks2021many}: renditions of ImageNet classes in the form of art, cartoons, and sculptures.
    \item \textbf{ImageNet-Sketch} (\textbf{IN-Sketch})~\cite{wang2019learning}: sketch-based depictions of ImageNet categories.
    \item \textbf{SI-Score}~\cite{djolonga2021robustness}: a synthetic benchmark that systematically varies object location, scale, and rotation.
\end{enumerate}
We use the standard ImageNet validation set (denoted \textbf{IN-V})~\cite{russakovsky2015imagenet} and ImageNet-v2 (denoted \textbf{IN-V2})~\cite{recht2019imagenet} as the in-distribution reference. For segmentation evaluation, we employ the ImageNet-Segmentation dataset~\cite{guillaumin2014imagenet}, which provides pixel-level masks for a subset of ImageNet classes.

\noindent\textbf{Baselines.}
We compare CFT against baselines that are most closely aligned with our goal: improve model robustness by modifying its saliency behavior. To ensure fairness, all methods are adapted to our fine-tuning setting:

\begin{enumerate}
    \item \textbf{GradMask}~\cite{simpson2019gradmask}: constrains model explanations to foreground regions via an input-gradient regularizer with human-annotated masks.
    \item \textbf{RRR}~\cite{ross2017right}: applies saliency-based gradient masking during backpropagation to reduce overfitting.
    \item \textbf{RobustViT}~\cite{chefer2022optimizing}: applies an efficient fine-tuning mechanism using uniform foreground-background segmentation masks guidance. We report both annotated and TokenCut segmentation versions (marked as annotated/TokenCut) of this method.
    \item \textbf{RRDA}~\cite{Santos_2023_ICCV}: employs explanation-guided data augmentation to preserve foreground relevance.
\end{enumerate}

\noindent Although these methods were originally designed to operate during full training, this is impractical for modern large-scale vision models due to the substantial computational cost. To ensure a fair evaluation, we integrate each baseline objective into a comparable fine-tuning procedure that mirrors our own setup. Notably, RobustViT is designed for ViT architectures; therefore, it is not evaluated on the CNv2 model.

\subsection{Training and Implementation Details}
\noindent\textbf{Training Procedure.} In most experiments, we follow the protocol of~\cite{yosinski2014transferable}, which examined transfer learning on half of the classes.
Specifically, we construct a small finetuning dataset by sampling three images per class for half of the classes in ImageNet-1K~\cite{deng2009imagenet}, totaling 1,500 images. This sparse sampling is computationally motivated and tests the method's data efficiency. We select classes randomly to ensure diverse semantic coverage. All models are initialized from publicly available, ImageNet-1K pre-trained checkpoints. Fine-tuning for all models is performed for 50 epochs using the AdamW optimizer with a batch size of 8. Learning rates are selected via grid search in $[5\!\times\!10^{-7}, 5\!\times\!10^{-6}]$ for every model. CFT uses fixed loss weights $\lambda_{\text{non-concept}}\!=\!1.2$, $\lambda_{\text{concept}}\!=\!0.5$, $\lambda_{\text{align}}\!=\!0.8$, and $\lambda_{\text{cls}}\!=\!0.2$ across all models and datasets.

\noindent\textbf{Concept Set Creation.} For concept set creation, we used $P=30$ samples for every class, guided by the feedback from the occurrence rate and spatial coverage measurements. This process was conducted using occurrence rate $\geq$15\% and spatial coverage $\geq$20\%, and resulted with a total of 1852 concepts over 500 classes (half of the classes in ImageNet-1K~\cite{deng2009imagenet}). In contrast to Oikarinen et al.~\cite{oikarinen2023label}, who use GPT-3 for concept set creation, we employ the GPT-4o-mini model~\cite{achiam2023gpt}, while keeping the remainder of the setup consistent with the original work.

\noindent All experiments are conducted on NVIDIA A100 GPUs using PyTorch. Code and reproducibility details are available in the Github repository. Further implementation details are provided in the Appendix.

\noindent Best results are in \textbf{bold}, second-best are \underline{underlined}.

\subsection{Results}
\begin{table}[t!]
\centering
\caption{Out-of-distribution (OOD) robustness and in-distribution accuracy: metrics are Top-1 and Top-5 accuracy (\%).}
\label{tab:main_robustness}
\resizebox{0.49\textwidth}{!}{
\begin{tabular}{llccccccc}
\toprule
\textbf{Dataset} & \textbf{Model} & \textbf{Metric} & \textbf{Original} & \textbf{GradMask} & \textbf{RRR} & \textbf{RRDA} & \textbf{RobustViT} & \textbf{CFT} \\
\midrule
\multirow{8}{*}{\textbf{IN-V}}
& \multirow{2}{*}{ViT-B} & R@1 & 80.41 & 80.65 & 80.89 & \underline{81.12} & 80.24/80.11 & \textbf{81.35} \\
& & R@5 & 94.88 & 95.03 & 95.17 & \underline{95.39} & 94.16/94.25 & \textbf{95.51} \\
\cmidrule(lr){2-9}
& \multirow{2}{*}{DINOv2} & R@1 & 81.07 & 80.95 & 80.48 & \textbf{81.61} & 80.38/80.72 & \underline{81.44} \\
& & R@5 & 95.20 & 95.08 & 94.83 & \underline{95.53} & 95.05/95.12 & \textbf{95.65} \\
\cmidrule(lr){2-9}
& \multirow{2}{*}{DeiT} & R@1 & 82.20 & 82.04 & 81.85 & \textbf{82.78} & 81.93/81.78 & \underline{82.61} \\
& & R@5 & 95.04 & 95.23 & 95.37 & \underline{95.64} & 94.82/94.89 & \textbf{95.77} \\
\cmidrule(lr){2-9}
& \multirow{2}{*}{CNv2} & R@1 & 86.49 & 86.68 & 86.74 & \underline{87.18} & - & \textbf{87.27} \\
& & R@5 & 95.05 & 95.24 & 95.09 & \underline{95.42} & - & \textbf{95.71} \\
\midrule
\multirow{8}{*}{\textbf{IN-V2}}
& \multirow{2}{*}{ViT-B} & R@1 & 68.32 & 68.51 & 68.68 & \underline{69.04} & 67.74/67.85 & \textbf{69.19} \\
& & R@5 & 83.74 & 83.95 & 84.18 & \underline{84.60} & 83.43/83.67 & \textbf{84.77} \\
\cmidrule(lr){2-9}
& \multirow{2}{*}{DINOv2} & R@1 & 71.32 & 71.55 & 71.45 & \textbf{72.08} & 71.03/71.29 & \underline{71.91} \\
& & R@5 & 87.42 & 87.61 & 87.53 & \underline{88.02} & 87.11/87.36 & \textbf{88.15} \\
\cmidrule(lr){2-9}
& \multirow{2}{*}{DeiT} & R@1 & 72.43 & 72.68 & 72.75 & \underline{72.85} & 72.68/72.81 & \textbf{73.11} \\
& & R@5 & 87.86 & 88.05 & 88.11 & \underline{88.35} & 87.61/87.89 & \textbf{88.58} \\
\cmidrule(lr){2-9}
& \multirow{2}{*}{CNv2} & R@1 & 74.49 & 74.70 & 74.83 & \underline{75.09} & - & \textbf{75.25} \\
& & R@5 & 88.77 & 88.98 & 88.93 & \underline{89.34} & - & \textbf{89.50} \\
\midrule
\multirow{8}{*}{\textbf{IN-A}}
& \multirow{2}{*}{ViT-B} & R@1 & 13.26 & 15.37 & 18.45 & \underline{25.16} & 24.88/23.81 & \textbf{27.76} \\
& & R@5 & 32.54 & 38.29 & 45.43 & \underline{59.87} & 59.28/56.61 & \textbf{62.75} \\
\cmidrule(lr){2-9}
& \multirow{2}{*}{DINOv2} & R@1 & 14.92 & 16.73 & 19.25 & \underline{25.74} & 25.22/23.86 & \textbf{27.71} \\
& & R@5 & 34.22 & 39.14 & 46.91 & \underline{60.01} & 58.76/57.31 & \textbf{62.36} \\
\cmidrule(lr){2-9}
& \multirow{2}{*}{DeiT} & R@1 & 15.37 & 17.65 & 19.94 & \underline{25.61} & 22.15/24.98 & \textbf{27.72} \\
& & R@5 & 35.43 & 41.08 & 47.92 & \underline{60.09} & 58.62/59.85 & \textbf{62.20} \\
\cmidrule(lr){2-9}
& \multirow{2}{*}{CNv2} & R@1 & 16.25 & 18.46 & 20.93 & \underline{25.89} & - & \textbf{27.93} \\
& & R@5 & 36.07 & 42.34 & 49.15 & \underline{59.88} & - & \textbf{62.40} \\
\midrule
\multirow{8}{*}{\textbf{ObjectNet}}
& \multirow{2}{*}{ViT-B} & R@1 & 33.26 & 36.44 & 40.32 & \underline{51.12} & 49.22/47.59 & \textbf{54.28} \\
& & R@5 & 50.54 & 56.72 & 62.18 & \underline{73.02} & 72.87/71.32 & \textbf{75.46} \\
\cmidrule(lr){2-9}
& \multirow{2}{*}{DINOv2} & R@1 & 34.93 & 38.27 & 42.51 & \underline{52.42} & 52.21/52.13 & \textbf{53.89} \\
& & R@5 & 51.81 & 58.02 & 63.17 & \underline{74.12} & 73.63/73.89 & \textbf{75.58} \\
\cmidrule(lr){2-9}
& \multirow{2}{*}{DeiT} & R@1 & 35.42 & 38.86 & 43.26 & \underline{52.81} & 45.01/46.38 & \textbf{54.24} \\
& & R@5 & 52.26 & 58.49 & 63.83 & \underline{74.03} & 69.23/69.84 & \textbf{75.46} \\
\cmidrule(lr){2-9}
& \multirow{2}{*}{CNv2} & R@1 & 36.18 & 39.43 & 43.82 & \underline{52.54} & - & \textbf{54.19} \\
& & R@5 & 53.01 & 59.04 & 64.21 & \underline{74.03} & - & \textbf{75.62} \\
\midrule
\multirow{8}{*}{\textbf{IN-R}}
& \multirow{2}{*}{ViT-B} & R@1 & 30.26 & 33.19 & 37.21 & \underline{47.12} & 45.93/44.28 & \textbf{48.47} \\
& & R@5 & 45.54 & 50.37 & 56.41 & \underline{69.02} & 68.71/67.33 & \textbf{70.50} \\
\cmidrule(lr){2-9}
& \multirow{2}{*}{DINOv2} & R@1 & 31.43 & 34.25 & 38.02 & \underline{47.93} & 46.84/47.17 & \textbf{48.53} \\
& & R@5 & 46.12 & 51.05 & 57.12 & \underline{70.12} & 69.43/69.87 & \textbf{70.73} \\
\cmidrule(lr){2-9}
& \multirow{2}{*}{DeiT} & R@1 & 32.14 & 35.06 & 38.75 & \underline{47.68} & 44.82/45.17 & \textbf{48.33} \\
& & R@5 & 46.77 & 51.63 & 57.41 & \underline{70.04} & 60.34/61.79 & \textbf{70.65} \\
\cmidrule(lr){2-9}
& \multirow{2}{*}{CNv2} & R@1 & 32.86 & 35.78 & 39.16 & \underline{47.72} & - & \textbf{48.37} \\
& & R@5 & 47.28 & 52.08 & 57.64 & \underline{70.03} & - & \textbf{70.68} \\
\midrule
\multirow{8}{*}{\textbf{IN-Sketch}}
& \multirow{2}{*}{ViT-B} & R@1 & 35.49 & 36.00 & 36.56 & \underline{36.85} & 36.78/36.62 & \textbf{37.06} \\
& & R@5 & 54.89 & 56.12 & 57.75 & \underline{61.94} & 61.82/61.46 & \textbf{62.59} \\
\cmidrule(lr){2-9}
& \multirow{2}{*}{DINOv2} & R@1 & 41.32 & 42.14 & 42.92 & \underline{44.50} & 43.62/44.08 & \textbf{44.74} \\
& & R@5 & 63.67 & 64.67 & 65.65 & \underline{68.61} & 68.22/68.51 & \textbf{68.90} \\
\cmidrule(lr){2-9}
& \multirow{2}{*}{DeiT} & R@1 & 42.43 & 43.09 & 43.61 & \underline{44.50} & 43.86/44.32 & \textbf{44.83} \\
& & R@5 & 64.84 & 65.71 & 66.52 & \underline{69.21} & 68.48/68.73 & \textbf{69.55} \\
\cmidrule(lr){2-9}
& \multirow{2}{*}{CNv2} & R@1 & 42.49 & 43.16 & 43.80 & \underline{45.87} & - & \textbf{46.14} \\
& & R@5 & 65.17 & 66.07 & 66.93 & \underline{70.52} & - & \textbf{70.81} \\
\bottomrule
\end{tabular}
}
\end{table}
\begin{table}[t]
\centering
\caption{Robustness to geometric transformations on the SI-Score benchmark: metrics are Top-1 and Top-5 accuracy (\%).}
\label{tab:si_score}
\resizebox{0.49\textwidth}{!}{
\begin{tabular}{l l l c c c c c c}
\toprule
\textbf{Dataset} & \textbf{Model} & \textbf{Metric} & \textbf{Original} & \textbf{GradMask} & \textbf{RRR} & \textbf{RRDA} & \textbf{RobustViT} & \textbf{CFT} \\
\midrule

\multirow{10}{*}{\textbf{SI-location}}
& \multirow{2}{*}{ViT-B} & R@1 & 34.26 & 35.18 & 36.42 & 38.54 & \underline{39.02}/38.51 & \textbf{39.15} \\
& & R@5 & 50.54 & 52.33 & 54.21 & 60.95 & \underline{61.74}/60.88 & \textbf{62.48} \\[4pt]
\cmidrule(lr){2-9}
& \multirow{2}{*}{DINOv2} & R@1 & 35.11 & 36.05 & 37.22 & \underline{38.95} & 38.79/38.21 & \textbf{39.07} \\
& & R@5 & 52.85 & 54.67 & 56.40 & \underline{61.33} & 61.18/60.75 & \textbf{62.15} \\[4pt]
\cmidrule(lr){2-9}
& \multirow{2}{*}{DeiT} & R@1 & 34.74 & 35.41 & 36.93 & \underline{38.79} & 38.34/38.13 & \textbf{39.08} \\
& & R@5 & 51.24 & 53.06 & 55.38 & \underline{60.65} & 60.37/59.81 & \textbf{62.04} \\[4pt]
\cmidrule(lr){2-9}
& \multirow{2}{*}{CNv2} & R@1 & 35.05 & 36.20 & 37.14 & \underline{38.93} & - & \textbf{39.00} \\
& & R@5 & 52.10 & 54.45 & 56.89 & \underline{61.27} & - & \textbf{62.11} \\

\midrule

\multirow{10}{*}{\textbf{SI-rotation}}
& \multirow{2}{*}{ViT-B} & R@1 & 40.26 & 42.15 & 44.83 & 50.73 & \underline{51.69}/50.28 & \textbf{52.15} \\
& & R@5 & 56.54 & 59.07 & 62.88 & 69.02 & \underline{70.35}/67.54 & \textbf{71.48} \\[4pt]
\cmidrule(lr){2-9}
& \multirow{2}{*}{DINOv2} & R@1 & 41.12 & 43.26 & 45.74 & \underline{50.35} & 49.58/48.21 & \textbf{51.92} \\
& & R@5 & 58.17 & 60.32 & 63.09 & \underline{69.83} & 68.12/66.39 & \textbf{70.94} \\[4pt]
\cmidrule(lr){2-9}
& \multirow{2}{*}{DeiT} & R@1 & 42.03 & 43.95 & 46.40 & \underline{50.27} & 49.74/48.23 & \textbf{51.64} \\
& & R@5 & 57.96 & 60.41 & 63.88 & \underline{69.10} & 68.28/67.79 & \textbf{70.66} \\[4pt]
\cmidrule(lr){2-9}
& \multirow{2}{*}{CNv2} & R@1 & 41.66 & 43.75 & 46.00 & \underline{50.11} & - & \textbf{51.57} \\
& & R@5 & 58.05 & 60.72 & 63.16 & \underline{69.04} & - & \textbf{70.45} \\

\midrule

\multirow{10}{*}{\textbf{SI-size}}
& \multirow{2}{*}{ViT-B} & R@1 & 55.26 & 57.45 & 59.14 & \underline{62.72} & 62.49/61.28 & \textbf{63.15} \\
& & R@5 & 74.54 & 77.02 & 80.19 & \underline{86.21} & 85.94/84.23 & \textbf{87.48} \\[4pt]
\cmidrule(lr){2-9}
& \multirow{2}{*}{DINOv2} & R@1 & 56.01 & 58.33 & 59.88 & \underline{62.85} & 61.72/61.24 & \textbf{63.02} \\
& & R@5 & 75.11 & 78.46 & 81.17 & \underline{86.03} & 85.38/84.86 & \textbf{87.12} \\[4pt]
\cmidrule(lr){2-9}
& \multirow{2}{*}{DeiT} & R@1 & 55.88 & 57.93 & 59.54 & \underline{62.66} & 61.54/60.64 & \textbf{63.07} \\
& & R@5 & 75.20 & 77.38 & 80.46 & \underline{86.11} & 85.43/84.59 & \textbf{87.24} \\[4pt]
\cmidrule(lr){2-9}
& \multirow{2}{*}{CNv2} & R@1 & 56.14 & 58.12 & 59.67 & \underline{62.91} & - & \textbf{63.10} \\
& & R@5 & 75.48 & 78.02 & 81.08 & \underline{86.25} & - & \textbf{87.36} \\

\bottomrule
\end{tabular}
} \vspace{-3mm}
\end{table}

\begin{figure}[t]
  \centering
  \includegraphics[width=0.46\textwidth]{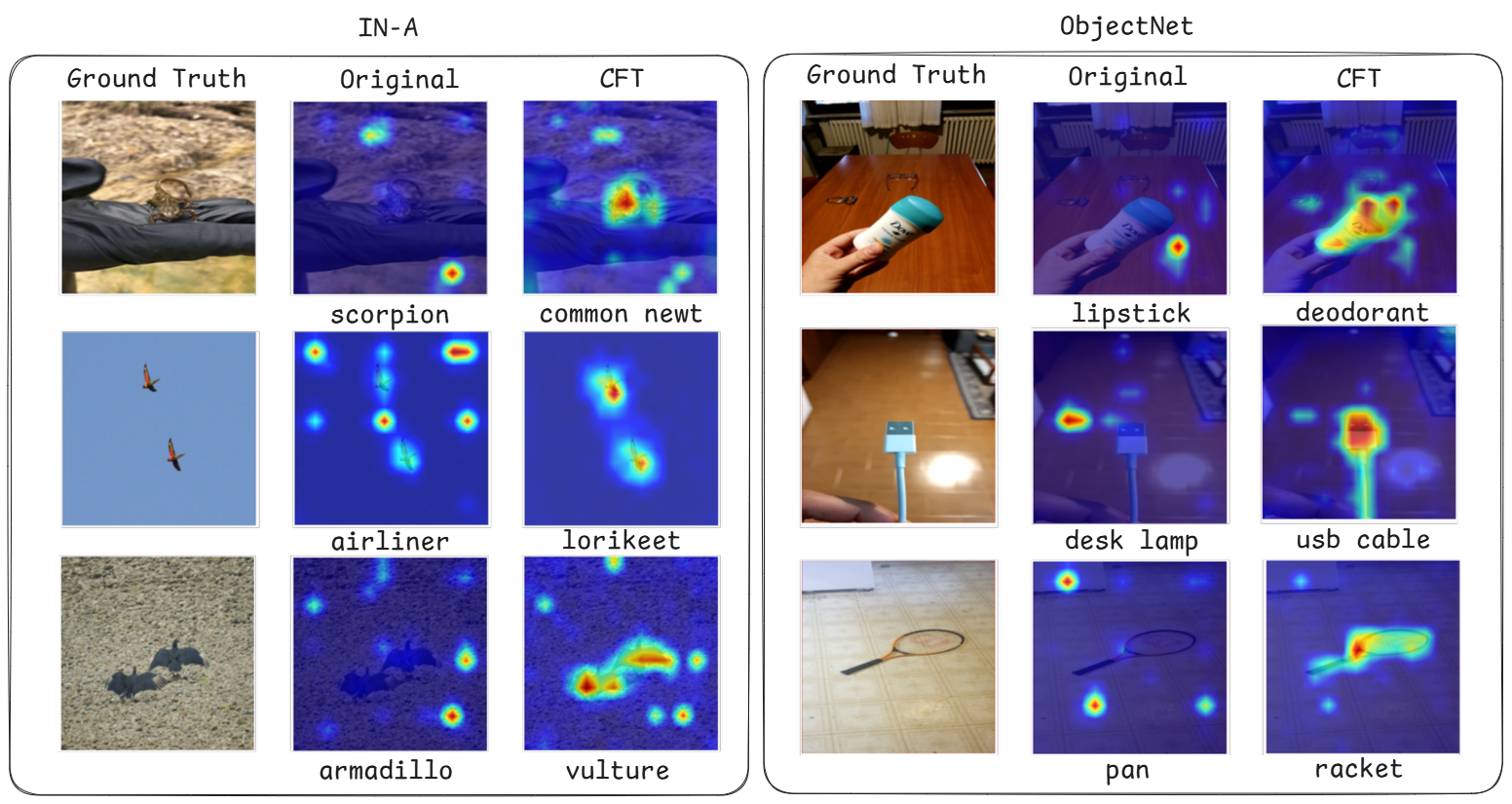}

\caption{Qualitative examples of CFT correcting prediction failures on OOD datasets using the ViT-B model: the baseline model (Original) misclassifies the images, with relevance maps often highlighting misleading context. Our CFT-finetuned model successfully corrects the prediction (e.g., ``scorpion'' $\rightarrow$ ``common newt'') by focusing its relevance on the object's core semantic concepts, demonstrating improved reasoning.}
  \label{fig:qualitive}
\end{figure}

Figure~\ref{fig:qualitive} presents qualitative examples from the IN-A and ObjectNet datasets using the ViT-B model, illustrating cases where CFT successfully corrected the model’s predictions. These examples vividly demonstrate the baseline model's failure mode: a strong reliance on spurious contextual cues. For instance, in the top row (IN-A), the baseline model misclassifies a ``common newt'' as a ``scorpion'', with its relevance map (Original) incorrectly diffusing across the textured background. After fine-tuning with CFT, the model not only corrects the prediction but also shifts its relevance to be tightly concentrated on the object's body. This provides qualitative evidence that CFT is successfully steering the model's reasoning from misleading cues toward the core object.

\noindent\textbf{Robustness under distribution shift.}
Table~\ref{tab:main_robustness} presents the Top-1 (\textbf{R@1}) and Top-5 (\textbf{R@5}) accuracies across different models and datasets. As shown, CFT consistently achieves substantial performance gains on real-world datasets, including adversarial variants (IN-A) and those featuring randomized or controlled backgrounds, rotations, and viewpoints (ObjectNet). In contrast, the improvement is less pronounced for datasets depicting artistic or abstract representations (IN-R, IN-Sketch), which often lack complex or varied backgrounds. This behavior is expected, as such datasets inherently minimize background biases. Furthermore, while baseline methods largely maintain their accuracy on datasets drawn from the original ImageNet distribution (IN-V and IN-V2), they exhibit clear degradation on real-world out-of-distribution datasets such as IN-A and ObjectNet. This observation suggests that existing methods are less effective at mitigating overfitting to the ImageNet domain.
Finally, we can observe that sometimes CFT introduces a minor reduction in accuracy on in-distribution data (IN-V and IN-V2), which can be reasonably interpreted as a result of improved regularization that alleviates overfitting to the training distribution.
On the synthetic SI-Score benchmark (Table~\ref{tab:si_score}), CFT demonstrates even more pronounced gains. This suggests that concept-focused reasoning inherently improves invariance to geometric transformations, as the model learns to rely on object structure and relevant features rather than absolute position or orientation cues.

\noindent\textbf{Relevance map alignment.} To verify that CFT indeed shifts model focus toward the object's relevant features and foreground information, we evaluate segmentation metrics on relevance maps (Table~\ref{tab:segmentation}). Using the ImageNet-Segmentation dataset~\cite{guillaumin2014imagenet}, we compute pixel accuracy \textbf{(PA)}, mean Intersection-over-Union \textbf{(mIoU)}, and mean Average Precision \textbf{(mAP)} between relevance maps and ground-truth masks. CFT improves all metrics across all architectures, confirming that our fine-tuning successfully aligns model explanations with object regions.

\begin{table}[t]
\centering
\small
\caption{Alignment of relevance maps with ground-truth object masks: pixel-level agreement between model relevance maps and human-annotated masks. Additional details are provided in Sec.~\ref{sec:experiments}.}
\label{tab:segmentation} 
\begin{tabular}{l l c c c c} 
\toprule
& Metric & ViT-B & DINOv2 & DeiT & CNv2 \\
\midrule

\multirow{3}{*}{Original} 
 & mIoU & 62.91 & 60.35 & 60.45 & 62.64 \\
 & mAP & 78.67 & 80.14 & 82.37 & 84.46 \\
 & PA   & 72.23 & 74.85 & 77.16 & 79.32 \\
\midrule
\multirow{3}{*}{CFT} 
 & mIoU & \textbf{68.23} & \textbf{70.84} & \textbf{72.91} & \textbf{74.21} \\
 & mAP  & \textbf{84.26} & \textbf{86.45} & \textbf{88.16} & \textbf{89.32} \\
 & PA   & \textbf{80.34} & \textbf{81.92} & \textbf{83.74} & \textbf{84.58} \\

\bottomrule
\end{tabular}
\end{table}

\noindent\textbf{Generalization across classes.} To verify that the robustness improvements induced by CFT fine-tuning extend beyond the classes used during training, we evaluate performance separately on training and non-training classes. Table~\ref{tab:train_vs_nontrain} reports the average improvement across models. The results indicate that both subsets achieve comparable gains on robustness benchmarks. As expected, classes included in the training set exhibit slightly higher accuracy on datasets derived from the original ImageNet distribution, reflecting their direct exposure during fine-tuning.

\begin{table*}[t]
\centering
\small
\caption{Generalization to unseen classes: robustness evaluation was performed separately on classes included in the fine-tuning set and those excluded from it. Last row reports average change for both training and non-training classes across all models and datasets.}
\label{tab:train_vs_nontrain}
\resizebox{0.98\textwidth}{!}{%
\begin{tabular}{ccc *{6}{cc}}
\toprule
\multirow{2}{*}{\textbf{Model}} & \multirow{2}{*}{\textbf{Train Classes}} & \multirow{2}{*}{\textbf{Method}} & \multicolumn{2}{c}{\textbf{IN-V}} & \multicolumn{2}{c}{\textbf{IN-A}} & \multicolumn{2}{c}{\textbf{IN-R}} & \multicolumn{2}{c}{\textbf{IN-Sketch}} & \multicolumn{2}{c}{\textbf{IN-V2}} & \multicolumn{2}{c}{\textbf{ObjectNet}} \\
\cmidrule(lr){4-5} \cmidrule(lr){6-7} \cmidrule(lr){8-9} \cmidrule(lr){10-11} \cmidrule(lr){12-13} \cmidrule(lr){14-15}
& & & R@1 & R@5 & R@1 & R@5 & R@1 & R@5 & R@1 & R@5 & R@1 & R@5 & R@1 & R@5 \\
\midrule
\multirow{4}{*}{\textbf{ViT-B}} & \multirow{2}{*}{$\checkmark$} & Original & 82.02 & 96.26 & 14.86 & 35.26 & 32.96 & 47.45 & 36.85 & 58.42 & 71.39 & 89.86 & 36.03 & 55.82 \\
& & CFT & \textbf{82.39} & \textbf{96.42} & \textbf{22.95} & \textbf{46.10} & \textbf{35.53} & \textbf{50.59} & \textbf{38.16} & \textbf{59.73} & \textbf{71.46} & \textbf{89.92} & \textbf{43.51} & \textbf{64.67} \\
\cmidrule(lr){2-15}
& \multirow{2}{*}{$\times$} & Original & 81.04 & 95.95 & 17.20 & 38.74 & 34.73 & 49.62 & 34.02 & 56.41 & \textbf{71.33} & \textbf{89.85} & 34.33 & 53.88 \\
& & CFT & \textbf{81.33} & \textbf{96.12} & \textbf{25.43} & \textbf{48.74} & \textbf{37.04} & \textbf{52.17} & \textbf{35.73} & \textbf{57.94} & 71.21 & 89.63 & \textbf{41.01} & \textbf{62.33} \\
\midrule
\multirow{4}{*}{\textbf{DINOv2}} & \multirow{2}{*}{$\checkmark$} & Original & \textbf{82.89} & 96.04 & 24.10 & 50.39 & 38.65 & 55.75 & 43.52 & 65.29 & 73.22 & 91.09 & 38.58 & 63.68 \\
& & CFT & 82.80 & \textbf{96.48} & \textbf{31.45} & \textbf{57.87} & \textbf{43.43} & \textbf{59.68} & \textbf{46.00} & \textbf{67.54} & \textbf{73.67} & \textbf{91.39} & \textbf{44.01} & \textbf{69.76} \\
\cmidrule(lr){2-15}
& \multirow{2}{*}{$\times$} & Original & 83.80 & 96.15 & 21.42 & 45.52 & 41.11 & 57.52 & 40.33 & 63.74 & 72.11 & 91.24 & 41.38 & 61.54 \\
& & CFT & \textbf{84.09} & \textbf{96.53} & \textbf{29.08} & \textbf{53.68} & \textbf{43.59} & \textbf{61.01} & \textbf{40.79} & \textbf{64.96} & \textbf{72.30} & \textbf{91.68} & \textbf{47.24} & \textbf{67.98} \\
\midrule
\multirow{4}{*}{\textbf{DeiT}} & \multirow{2}{*}{$\checkmark$} & Original & 82.82 & 95.89 & 24.57 & 50.85 & 38.70 & 55.47 & 43.66 & 65.44 & 73.30 & 90.87 & 38.38 & 64.06 \\
& & CFT & \textbf{83.58} & \textbf{96.22} & \textbf{30.55} & \textbf{57.52} & \textbf{43.63} & \textbf{59.69} & \textbf{46.22} & \textbf{67.20} & \textbf{73.52} & \textbf{91.28} & \textbf{43.55} & \textbf{69.38} \\
\cmidrule(lr){2-15}
& \multirow{2}{*}{$\times$} & Original & 84.21 & 95.79 & 20.96 & 45.85 & 41.60 & 57.07 & 40.36 & 63.42 & \textbf{72.13} & \textbf{91.62} & 40.88 & 61.55 \\
& & CFT & \textbf{84.34} & \textbf{96.17} & \textbf{28.63} & \textbf{53.57} & \textbf{43.69} & \textbf{61.38} & \textbf{40.37} & \textbf{65.21} & 72.01 & 91.55 & \textbf{47.73} & \textbf{67.61} \\
\midrule
\multirow{4}{*}{\textbf{CNv2}} & \multirow{2}{*}{$\checkmark$} & Original & 83.82 & 97.29 & 25.32 & 51.64 & 39.84 & 56.97 & 44.75 & 66.52 & 74.45 & 92.35 & 39.81 & 64.91 \\
& & CFT & \textbf{84.28} & \textbf{97.56} & \textbf{33.12} & \textbf{59.87} & \textbf{45.36} & \textbf{61.66} & \textbf{47.93} & \textbf{69.48} & \textbf{75.63} & \textbf{92.90} & \textbf{45.96} & \textbf{71.65} \\
\cmidrule(lr){2-15}
& \multirow{2}{*}{$\times$} & Original & 85.03 & 97.35 & 22.68 & 46.78 & 42.35 & 58.74 & 41.59 & 64.94 & 73.34 & 92.49 & 42.61 & 62.77 \\
& & CFT & \textbf{85.70} & \textbf{97.48} & \textbf{31.04} & \textbf{55.63} & \textbf{45.55} & \textbf{62.95} & \textbf{42.72} & \textbf{66.88} & \textbf{73.69} & \textbf{92.83} & \textbf{49.17} & \textbf{69.96} \\
\midrule
\multicolumn{3}{c}{\textbf{Avg. Change ($\checkmark$)}} & \textbf{\color{green}+0.38} & \textbf{\color{green}+0.30} & \textbf{\color{green}+7.31} & \textbf{\color{green}+8.31} & \textbf{\color{green}+4.45} & \textbf{\color{green}+4.00} & \textbf{\color{green}+2.38} & \textbf{\color{green}+2.07} & \textbf{\color{green}+0.48} & \textbf{\color{green}+0.33} & \textbf{\color{green}+6.06} & \textbf{\color{green}+6.75} \\
\cmidrule(lr){4-15}
\multicolumn{3}{c}{\textbf{Avg. Change ($\times$)}} & \textbf{\color{green}+0.35} & \textbf{\color{green}+0.27} & \textbf{\color{green}+7.98} & \textbf{\color{green}+8.68} & \textbf{\color{green}+2.52} & \textbf{\color{green}+3.64} & \textbf{\color{green}+0.83} & \textbf{\color{green}+1.62} & \textbf{\color{green}+0.08} & \textbf{\color{green}+0.12} & \textbf{\color{green}+6.49} & \textbf{\color{green}+7.04} \\
\bottomrule
\end{tabular}
}
\end{table*}

\noindent In summary, CFT consistently enhances model robustness across architectures and distribution shifts by explicitly guiding relevance toward concepts and object foregrounds. Its gains generalize to unseen classes and are most pronounced in scenarios where background cues are misleading, a common failure mode in real-world deployment~\cite{geirhos2020shortcut}.

\subsection{Ablation Studies}
\label{subsec:ablation}
\begin{table*}[ht!]
\centering
\small
\caption{Ablation study: Concept-level guidance vs. object-level guidance.}
\label{tab:ablation_seg} 
\resizebox{0.96\textwidth}{!}{%
\begin{tabular}{ll c c c c c c c c c}
\toprule
\textbf{Model} & \textbf{Method} & \textbf{IN-V} & \textbf{IN-A} & \textbf{IN-R} & \textbf{IN-Sketch} & \textbf{IN-V2} & \textbf{ObjectNet} & \textbf{SI-location} & \textbf{SI-rotation} & \textbf{SI-size} \\
\midrule
\multirow{3}{*}{ViT-B}
& Original & \underline{81.53} & 16.09 & 33.81 & 35.47 & \underline{71.13} & 35.12 & 33.36 & 39.18 & 55.65 \\
& CFT & \textbf{82.57} & \textbf{26.05} & \textbf{36.69} & \textbf{36.28} & \textbf{71.65} & \textbf{42.29} & \textbf{38.60} & \textbf{46.82} & \textbf{61.94} \\
& Segmentation & 80.25 & \underline{24.12} & \underline{35.95} & \underline{36.01} & 70.84 & \underline{41.70} & \underline{38.13} & \underline{46.31} & \underline{61.40} \\
\midrule
\multirow{3}{*}{DINOv2}
& Original & \underline{83.15} & 17.52 & 35.20 & 36.83 & \underline{72.81} & 36.78 & 34.63 & 40.25 & 57.09 \\
& CFT & \textbf{83.62} & \textbf{27.88} & \textbf{38.02} & \textbf{37.61} & \textbf{73.04} & \textbf{43.91} & \textbf{39.82} & \textbf{48.19} & \textbf{63.11} \\
& Segmentation & 82.04 & \underline{25.91} & \underline{37.14} & \underline{37.12} & 72.25 & \underline{43.15} & \underline{39.07} & \underline{47.73} & \underline{62.85} \\
\bottomrule
\end{tabular}
}
\end{table*}
\begin{table*}[h!]
\centering
\small
\caption{Ablation study on loss components: we evaluate the impact of removing each of our three main loss terms using the ViT-B model.}
\label{tab:ablation_loss}
\resizebox{0.95\textwidth}{!}{%
\begin{tabular}{l c c c c c c c c c}
\toprule
\textbf{Method} & \textbf{IN-V} & \textbf{IN-A} & \textbf{IN-R} & \textbf{IN-Sketch} & \textbf{IN-V2} & \textbf{ObjectNet} & \textbf{SI-location} & \textbf{SI-rotation} & \textbf{SI-size} \\
\midrule
Original & 81.53 & 16.09 & 33.81 & 35.47 & 71.15 & 35.12 & 33.36 & 39.18 & 55.65 \\
CFT & 82.57 & \textbf{24.11} & \textbf{36.34} & \textbf{36.28} & 72.03 & \textbf{42.29} & \textbf{38.60} & \textbf{46.25} & \textbf{62.18} \\
w/o $\mathcal{L}_{\text{cls}}$ (Eq.~\ref{eq:cls_loss}) & 79.82 & 17.96 & 34.29 & 34.84 & 69.41 & 39.65 & 37.27 & 43.09 & 58.46 \\
w/o $\mathcal{L}_{\text{non-concept}}$ (Eq.~\ref{eq:non_concept_loss}) & 82.50 & 19.14 & 35.13 & 36.26 & 72.01 & 41.70 & 34.93 & 42.31 & 58.40 \\
w/o $\mathcal{L}_{\text{concept}}$ (Eq.~\ref{eq:concept_loss}) & 81.26 & 24.10 & 34.37 & 35.22 & 71.69 & 42.14 & 38.55 & 45.61 & 61.79 \\
w/ ground-truth & \textbf{82.61} & 21.35 & 34.56 & 35.72 & \textbf{72.08} & 41.55 & 37.02 & 44.68 & 61.36 \\
\bottomrule
\end{tabular}
}
\end{table*}

In what follows, we present three sets of experiments:
\textbf{(1)} a comparison between concept-based and object-based segmentation,
\textbf{(2)} an ablation study on the loss terms of the training objective, and
\textbf{(3)} an evaluation of different saliency methods for generating relevance maps.
These results demonstrate the clear advantage of using AttnLRP as the explanation method for CFT compared with alternative approaches.

\noindent Table~\ref{tab:ablation_seg} compares Top-1 accuracy (\%) results of concept-based guidance during fine-tuning (\textbf{CFT}) with object-segmentation–based guidance (\textbf{Segmentation}) for ViT-B and DINOv2 across all datasets. For this evaluation, we use the same loss function as in CFT, but replace the concept segmentation map $S(I)$ with the ground-truth object segmentation mask. Notably, we also experimented with GroundedSAM~\cite{ren2024groundingsam} by using the class label as a prompt and using the response mask instead of the ground-truth object mask. This approach produced results nearly identical to the Similarity baseline, maintaining the same performance trends. All experiments follow the same training setup as described previously. The goal of this experiment is to assess whether fine-grained semantic concepts provide a superior guidance signal for robustness than uniform object segments. As shown in the table, CFT consistently outperforms Segmentation guidance across both in-distribution and out-of-distribution datasets. This highlights the advantage of leveraging concept-based cues to enhance model robustness and, in some cases, improve in-distribution accuracy. While object-segmentation maps provide a reasonable level of robustness, using concept-guided masks further improves performance on standard in-distribution data.

\noindent Table~\ref{tab:ablation_loss} presents the ablation  results for the $\lambda_{\text{non-concept}}$, $\lambda_{\text{concept}}$, and $\lambda_{\text{cls}}$ loss terms using the ViT-B model across all datasets, evaluated by Top-1 accuracy (\%). Moreover, we conduct an ablation to assess the role of our classification-consistency classification loss (Eq.~\ref{eq:cls_loss}) by substituting it with the standard cross-entropy loss computed using the ground-truth label. Results show that performance on IN-V and IN-V2 is relatively insensitive to the removal of $\lambda_{\text{non-concept}}$. In contrast, this term plays a crucial role in out-of-distribution datasets, as its absence leads to a significant accuracy drop. Furthermore, $\lambda_{\text{cls}}$ proves essential for maintaining robustness, as its removal results in substantial performance degradation.

\noindent Finally, our ablation study highlights the advantage of employing the classification-consistency loss over the standard ground-truth cross-entropy. While the ground-truth variant maintains slightly higher original accuracy, the classification-consistency loss consistently yields greater improvements in model robustness.

\noindent Table~\ref{tab:abl1} reports CFT Top-1 accuracy (\%) performance using alternative relevance methods: Gradient-Rollout~\cite{iia}, IIA~\cite{iia}, and GradCAM~\cite{chefer2021transformer}. We evaluated ViT-B, on IN-A and IN-R. Across all evaluations, AttnLRP yields superior performance, highlighting its effectiveness for relevance propagation in the CFT approach.
\begin{table}[t!]
\centering
\caption{Ablation study on relevance methods: Top-1 accuracy using different relevance methods.}
\label{tab:abl1}
\scalebox{0.9}{
\begin{tabular}{l|cccc}
\toprule
\textbf{Method} & GradCAM & Gradient-Rollout & IIA & AttnLRP \\
\midrule
IN-A & 25.88 & 26.43 & \underline{26.94} & \textbf{27.82} \\
IN-R & 43.32 & 46.92 & \underline{47.58} & \textbf{48.54} \\
\bottomrule
\end{tabular}}
\end{table}
\section{Conclusion}
\label{sec:conclusion}
We introduced Concept-Guided Fine-Tuning (CFT), a fully automated framework designed to address a key limitation of modern vision models: their reliance on spurious correlations for classification. By steering the model’s internal reasoning away from such cues and toward semantically meaningful concepts, CFT substantially improves OOD robustness. Extensive experiments across five OOD benchmarks demonstrate that CFT consistently outperforms prior saliency-regularization approaches. 
Importantly, the robustness improvements generalize to classes that are not observed during fine-tuning, indicating that CFT promotes a more robust reasoning process rather than merely replacing one set of cues with another. Our ablation studies further support a central hypothesis: fine-grained semantic concepts provide a significantly stronger supervision signal for robustness than conventional foreground–background segmentation masks. Overall, CFT offers a scalable and interpretable pathway toward more reliable vision models. Limitations and future work are discussed in the Appendix.

\section*{Acknowledgment}
This research was supported by The Open University of Israel Research Fund.

{
    \small
    \bibliographystyle{ieeenat_fullname}
    \bibliography{main}
}

\clearpage
\setcounter{page}{1}
\maketitlesupplementary

\section{Implementation Details}
\label{sec:implementation}

\paragraph{Hyperparameters.}
In Table~\ref{tab:hyperparams}, we summarize the hyperparameter configurations used across all experiments. Our method is highly stable and relies on a consistent set of hyperparameters, with the exception of the learning rate. In contrast, RRR requires model-specific hyperparameter tuning and is notably sensitive to these choices. GradMask also demands careful learning rate tuning for each model. Its performance varies substantially with small adjustments to this parameter, and achieving convergence of the background loss proved challenging in both cases. RRDA shares no common hyperparameters with the other methods aside from the learning rate. CFT uses fixed loss weights $\lambda_{\text{non-concept}}\!=\!1.2$, $\lambda_{\text{concept}}\!=\!0.5$, $\lambda_{\text{align}}\!=\!0.8$, and $\lambda_{\text{cls}}\!=\!0.2$ across all models and datasets. We heavily weight the ${L}_{\text{non-concept}}$ loss, as reliance on spurious cues (areas without concepts in this case) is the primary issue to be corrected.

\begin{table*}[t]
\centering
\small
\caption{Hyperparameter selection for all methods.}
\label{tab:hyperparams}
\begin{tabular}{l l c c c c c} 
\toprule
& Model & $\lambda_{\text{align}}$ & $\lambda_{\text{cls}}$ & $\lambda_{\text{non-concept}}$ & $\lambda_{\text{concept}}$ & Learning rate \\
\midrule

\multirow{4}{*}{CFT} 
 & ViT-B & 0.8 & 0.2 & 1.2 & 0.5 & 5e$-$7 \\
 & DINOv2 & 0.8 & 0.2 & 1.2 & 0.5 & 6e$-$7 \\
 & DeiT   & 0.8 & 0.2 & 1.2 & 0.5 & 8e$-$7 \\
 & CNv2   & 0.8 & 0.2 & 1.2 & 0.5 & 3e$-$6 \\
\midrule
\multirow{4}{*}{RRR} 
 & ViT-B & - & 2e$-$6 & 1e$-$10 & - & 2e$-$6 \\
 & DINOv2 & - & 2e$-$8 & 1e$-$10 & - & 1e$-$5 \\
 & DeiT   & - & 2e$-$6 & 1e$-$10 & - & 5e$-$6 \\
 & CNv2   & - & 2e$-$6 & 1e$-$8 & - & 3e$-$6 \\
 \midrule
\multirow{4}{*}{GradMask} 
 & ViT-B & - & 0.1 & 50 & - & 0.001 \\
 & DINOv2 & - & 0.1 & 50 & - & 0.005 \\
 & DeiT   & - & 0.1 & 50 & - & 0.001 \\
 & CNv2   & - & 0.1 & 50 & - & 0.05 \\
 \midrule
\multirow{4}{*}{RRDA} 
 & ViT-B & - & - & - & - & 2e-6 \\
 & DINOv2 & - & - & - & - & 1e-5 \\
 & DeiT   & - & - & - & - & 5e-6 \\
 & CNv2   & - & - & - & - & 3e-6 \\

\bottomrule
\end{tabular}
\end{table*}

\paragraph{Ablation on Concept Validation Thresholds.}

Following the initial label-free concept discovery procedure of~\cite{oikarinen2023label}, we further refined the resulting pool to obtain a high-quality concept set. To this end, we enforced minimum thresholds of an occurrence rate of at least 15\% and spatial coverage of at least 20\%. Applying these criteria to IN produced 1{,}852 validated concepts. Across the dataset, concepts appeared in 29\% of images on average, and those satisfying the filtering criteria covered roughly 35\% of the relevant region. We examined the effect of varying the occurrence-rate and spatial-coverage thresholds on Top-1 accuracy for ViT-B evaluated on IN-A and IN-R. The best results were obtained using our default thresholds of 15\% and 20\%. Increasing the thresholds to 40\%/40\% reduced the number of concepts to 694 and led to a noticeable drop in performance (IN-A: 24.59, IN-R: 44.23), presumably because many informative concepts were discarded. Relaxing the thresholds to 5\%/10\% increased the concept count to 2{,}435 but introduced substantial noise, which similarly harmed performance (IN-A: 25.13, IN-R: 44.92).

\paragraph{Concept Set Creation.} 
For concept set construction, we used $P=30$ samples per class, guided by the occurrence rate and spatial coverage feedback. Using thresholds of occurrence rate $\geq 15\%$ and spatial coverage $\geq 20\%$, this process yielded a total of 1852 concepts across 500 classes (half of ImageNet-1K~\cite{deng2009imagenet}). The filtering (occurrence rate and spatial coverage feedback) proceeded in two stages: we first applied the occurrence-rate threshold, and then evaluated the remaining candidates using the spatial-coverage criterion. In our experiments, all concepts that passed the occurrence-rate filter also satisfied the 20\% coverage threshold. While not required for our study, this procedure could be extended with an iterative refinement step — potentially assisted by an LLM to identify additional concepts that jointly satisfy both constraints.


\paragraph{Clarification on the $P$ parameter.}
The parameter $P$ is used exclusively during the validation of the initial concept sets. We first generate the initial concept sets using the procedure of Oikarinen et al.~\cite{oikarinen2023label}. Then, for each class, we examine $P=30$ images to compute the occurrence rate and spatial coverage. These measurements are subsequently used to filter and refine the initial concept sets.

\section{Concept Validation Effect}
\label{sec:ablation_studies}



\paragraph{Effect of the optional concept validation step.}
We further compared CFT performance with and without the concept validation stage, evaluating Top-1 accuracy on both IN-A and IN-R. Without validation, CFT achieves 26.01 on IN-A (vs.\ 27.92 with validation) and 47.19 on IN-R (vs.\ 48.51 with validation). Although the validation step provides a consistent performance boost, the non-validated variant remains competitive and continues to outperform several robustness-oriented baselines (Tab.~\ref{tab:main_robustness}). Yet, using the validation step provides state-of-the-art performance, outperforming all other approaches.


\paragraph{Ablation on Concept Validation Thresholds.} Following the initial label-free concept discovery procedure of~\cite{oikarinen2023label}, we further refined the resulting pool to obtain a high-quality concept set. To this end, we enforced minimum thresholds of an occurrence rate of at least 15\% and spatial coverage of at least 20\%. Applying these criteria to IN produced 1{,}852 validated concepts. Across the dataset, concepts appeared in 29\% of images on average, and those satisfying the filtering criteria covered roughly 35\% of the relevant region. We examined the effect of varying the occurrence-rate and spatial-coverage thresholds on Top-1 accuracy for ViT-B evaluated on IN-A and IN-R. The best results were obtained using our default thresholds of 15\% and 20\%. Increasing the thresholds to 40\%/40\% reduced the number of concepts to 694 and led to a noticeable drop in performance (IN-A: 24.59, IN-R: 44.23), presumably because many informative concepts were discarded. Relaxing the thresholds to 5\%/10\% increased the concept count to 2{,}435 but introduced substantial noise, which similarly harmed performance (IN-A: 25.13, IN-R: 44.92).

\section{Main evaluation - full results}
The results in Table~\ref{tab:main_robustness} are averaged over five random seeds, where the subset of ImageNet classes used for fine-tuning is varied while keeping all other parameters fixed. Table~\ref{tab:main_robustness_full} reports the corresponding standard deviations for this experiment.

\begin{table*}[t!]
\centering
\caption{Evaluation over 5 different seeds.}
\label{tab:main_robustness_full}
\resizebox{\textwidth}{!}{
\begin{tabular}{llcccccc}
\toprule
Model & Metric & IN-V & IN-V2 & IN-A & ObjectNet & IN-R & IN-Sketch \\
\midrule
\multirow{2}{*}{ViT-B} 
& R@1 & 81.35 $\pm$0.28 & 69.19 $\pm$0.51 & 27.76 $\pm$0.14 & 54.28 $\pm$0.82 & 48.47 $\pm$0.39 & 37.06 $\pm$0.66 \\
& R@5 & 95.51 $\pm$0.47 & 84.77 $\pm$0.19 & 62.75 $\pm$0.73 & 75.46 $\pm$0.32 & 70.50 $\pm$0.56 & 62.59 $\pm$0.21 \\
\midrule
\multirow{2}{*}{DINOv2} 
& R@1 & 81.44 $\pm$0.61 & 71.91 $\pm$0.34 & 27.71 $\pm$0.80 & 53.89 $\pm$0.17 & 48.53 $\pm$0.54 & 44.74 $\pm$0.42 \\
& R@5 & 95.65 $\pm$0.42 & 88.15 $\pm$0.77 & 62.36 $\pm$0.54 & 75.58 $\pm$0.59 & 70.73 $\pm$0.18 & 68.90 $\pm$0.69 \\
\midrule
\multirow{2}{*}{DeiT} 
& R@1 & 82.61 $\pm$0.44 & 73.11 $\pm$0.34 & 27.72 $\pm$0.57 & 54.24 $\pm$0.71 & 48.33 $\pm$0.23 & 44.83 $\pm$0.36 \\
& R@5 & 95.77 $\pm$0.68 & 88.58 $\pm$0.49 & 62.20 $\pm$0.31 & 75.46 $\pm$0.15 & 70.65 $\pm$0.75 & 69.55 $\pm$0.53 \\
\midrule
\multirow{2}{*}{CNv2} 
& R@1 & 87.27 $\pm$0.43 & 75.25 $\pm$0.63 & 27.93 $\pm$0.41 & 54.19 $\pm$0.50 & 48.37 $\pm$0.78 & 46.14 $\pm$0.27 \\
& R@5 & 95.71 $\pm$0.52 & 89.50 $\pm$0.38 & 62.40 $\pm$0.65 & 75.62 $\pm$0.22 & 70.68 $\pm$0.46 & 70.81 $\pm$0.60 \\
\bottomrule
\end{tabular}
}
\end{table*}
\section{Limitations and Future Work}
\label{sec:limitations}

\subsection{Failure Cases}
\label{subsec:failure_cases}

Despite strong overall performance, CFT exhibits identifiable failure modes:
 \paragraph{Abstract or non-visual concepts.}
        GPT-4o-mini rarely generates concepts that are semantically
        appropriate but not visually grounded (e.g., ``aggressive behaviour''
        for a lion). GroundedSAM cannot localize such concepts, resulting in empty
        high-confidence masks.
    \paragraph{Very small object parts.}
        For parts occupying $<\!2\%$ of image area (e.g., the beak of a
        distant bird), GroundedSAM's hit rate decreases. The impact
        on final accuracy is limited, as the remaining concepts provide
        sufficient coverage, but fine-grained part-level reasoning may be
        impaired.
\paragraph{Domain mismatch between LLM and target domain.}
        In specialized domains (medical imaging, satellite imagery),
        GPT-4o-mini's concept vocabularies may be imprecise or incomplete.
        In such settings, domain-specific LLMs or expert-curated concept
        lists are recommended.

\subsection{Limitations}
While our proposed CFT framework demonstrates improvements in model robustness across multiple benchmarks, several limitations warrant discussion.

\paragraph{Dependency on Vision-Language Models.} Our approach relies on the quality and capabilities of GroundedSAM for concept localization. While this eliminates the need for manual annotations, it introduces a dependency on the grounding model's performance. In cases where GroundedSAM fails to accurately segment concepts, particularly for abstract or fine-grained semantic attributes, the quality of guidance masks may degrade, potentially limiting CFT's effectiveness.

\paragraph{Computational Overhead.} While CFT is designed as a lightweight fine-tuning procedure requiring only 1,500 images, the initial concept creation and validation stage involves processing 30 samples per class through GroundedSAM, which introduces non-negligible computational costs. For datasets with thousands of classes, this preprocessing step could become a practical bottleneck. Moreover, computing relevance maps via AttnLRP during training adds overhead compared to standard gradient-based methods, though this cost is amortized across the fine-tuning procedure.

\paragraph{Architecture Specificity.} Although we demonstrate CFT's applicability to both ViTs and CNNs (ConvNeXt-V2), the primary design and optimization were conducted with transformer architectures in mind. The adaptation to CNNs, while successful, required modifications to the relevance computation procedure. Extending CFT to other emerging architectures may require additional architectural considerations.


\subsection{Future Work}

Several promising directions emerge from this work that could further advance concept-guided robustness in vision models.

\paragraph{Adaptive Concept Weighting.} Our current approach treats all validated concepts equally during fine-tuning. However, different concepts may contribute unequally to robustness for specific distribution shifts. Developing methods to dynamically weight concepts based on their discriminative power or relevance to particular OOD scenarios could yield more targeted robustness improvements. This could be achieved through simple masking response-based approaches or by using concept activation vectors (CAVs) to produce concept-class importance weights.

\paragraph{Hierarchical and Compositional Concepts.} Our framework currently treats concepts as independent entities. However, real-world objects exhibit hierarchical structure and compositional semantics. Incorporating compositional reasoning, where complex concepts are built from simpler primitives, could enhance both interpretability and robustness. 

\paragraph{Application to Other Domains.} Although this work focuses on image classification, the underlying principle of aligning model reasoning with semantically meaningful concepts extends naturally to other computer vision tasks (e.g., object detection, semantic segmentation, video understanding) and potentially to non-vision domains where structured, interpretable representations are valuable. Exploring these extensions could validate the generality of concept-guided learning as a robustness paradigm.

\end{document}